\documentclass{article}
\pdfoutput=1
\usepackage[preprint, nonatbib]{nips_2018}

\usepackage[utf8]{inputenc} % allow utf-8 input
\usepackage[T1]{fontenc}    % use 8-bit T1 fonts
\usepackage{hyperref}       % hyperlinks
\usepackage{url}            % simple URL typesetting
\usepackage{booktabs}       % professional-quality tables
\usepackage{amsfonts}       % blackboard math symbols
\usepackage{nicefrac}       % compact symbols for 1/2, etc.
\usepackage{microtype}      % microtypography

\usepackage{amsmath}
\usepackage{graphicx}
\usepackage{float}
\usepackage[caption = false]{subfig}
\usepackage{longtable}
\usepackage{booktabs}
\usepackage{multirow}
\usepackage{siunitx}
\usepackage{algorithm}
\usepackage[noend]{algpseudocode}
\usepackage{mathpazo}

\title{Deep Q learning for fooling neural networks}

\author{
  Mandar Kulkarni\\
  Data Scientist\\
  Schlumberger\\
  \texttt{mkulkarni9@slb.com} \\
  }

\begin{document}

\maketitle

\begin{abstract}
Deep learning models are vulnerable to external attacks.
In this paper, we propose a Reinforcement Learning (RL) based approach to generate adversarial examples for the pre-trained (target) models. We assume a semi black-box setting where the only access an adversary has to the target model is the class probabilities obtained for the input queries.
We train a Deep Q Network (DQN) agent which, with experience, learns to attack only a small portion of image pixels to generate non-targeted adversarial images. Initially, an agent explores an environment by sequentially modifying random sets of image pixels and observes its effect on the class probabilities. At the end of an episode, it receives a positive (negative) reward if it succeeds (fails) to alter the label of the image. Experimental results with MNIST, CIFAR-10 and Imagenet datasets demonstrate that our RL framework is able to learn an effective attack policy.    
\end{abstract}

\section{Introduction and related works}

Deep learning models are vulnerable to external attacks such as adversarial inputs. The examples from the dataset can be perturbed in a manner that a human assigns the same label to it, however, it forces machine learning models to mis-classify it. They pose a potential threat to machine learning models when deployed in the real world \cite{kurakin2016adversarial}. 
Su et al. \cite{su2017one} proposed a differential evolution based approach which generates the adversarial examples by modifying a single pixel in the input image. We suspect that different regions in the image have variable sensitivity to adversarial attack and we attempt to exploit it to generate adversarial images with minimal changes to the input image.   

\begin{figure} [!t]
\centering
\begin{tabular}{c c}

\includegraphics[width=300pt, height = 120pt]{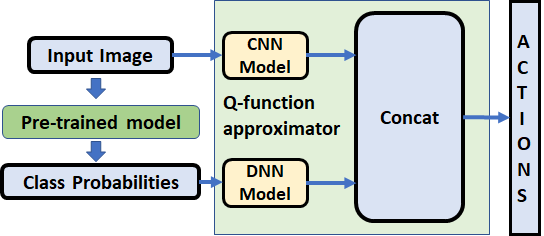}&
\includegraphics[width=110pt, height = 110pt]{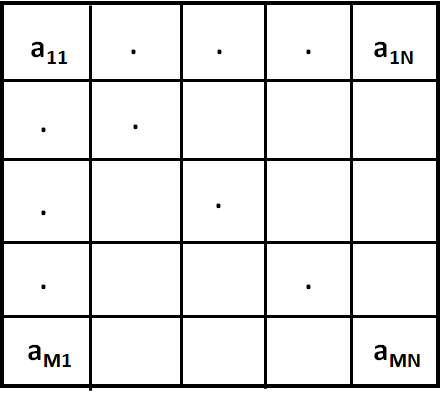}\\(a)&(b)\\

\end{tabular}
\caption{\label{fig:inpt117} Details of DQN approach (a) input image is represented by the CNN while the class probability vector is represented by the DNN, (b) possible actions for an agent.}
\end{figure}

In this paper, we propose a Deep reinforcement learning approach which eventually learns to perform non-targeted adversarial attacks by modifying a small portion (0.5-1\%) of input image pixels. 
We assume a semi black-box setting where the only access an adversary (RL agent) has to the target (pre-trained) model is the class probabilities obtained for the query image. 
RL agent does not have any knowledge of target model's architecture or gradients which makes it agnostic to the specific target function. 
We assume that the state of the environment can be inferred from the input image and its class probability vector.
The action space $A$ of an agent consists of set of small non-overlapping blocks in the input image (Fig. \ref{fig:inpt117}(b)). 
At each time step, an agent chose an action to modify a block of the image and queries the target model to obtain the class probabilities. In an episode, the agent sequentially modifies different blocks in the image and obtains a sequence of perturbed images and corresponding class probability vectors.
If the agent succeeds in modifying the label of the image within fixed time steps, it receives a strong positive rewards else, it gets a negative reward.
We train a Deep Q Network (DQN) \cite{mnih2013playing} \cite{li2017deep} on these episodes to learn an effective attack policy. 

We experimented with models trained on MNIST and CIFAR-10. We also demonstrate early results on the DenseNet-121 \cite{huang2017densely} model trained on Imagenet. Results indicate that, with minimal changes to the input image, the RL agent is able to fool the target models. The python code for experiments with MNIST and CIFAR-10 datasets is available at \url{https://github.com/mandareln/deep-q-learning-adversarial}

\section{Methodology}
Our adversarial RL agent learns the attack policy
by interacting with the environment. We assume that the state of the environment is observable through the input image and its corresponding class probabilities obtained from the target model. The state at time $t$ is represented as  $s_t \in (x_t, p_t)$, where $x_t$ and $p_t$ denotes the input image and the class probabilities, respectively. 
At each time step, an agent performs an action $a_t \in a_{ij} (i \in [1,M], j\in [1,N]$) to modify a single block of image as follows 
\begin{eqnarray}\label{eq:so1}
x_{t+1}[a_{ij}] = x_t[a_{ij}] + \lambda
\end{eqnarray}
where $\lambda$ denotes the change magnitude.
Fig. \ref{fig:inpt117}(b) shows the possible action space of the agent.
Agent queries the target model with the modified image and obtains the updated class probabilities. The next state of the environment is thus represented as $(x_{t+1},p_{t+1})$.
$M$ and $N$ denotes the number of blocks along the row and column respectively. Therefore, for each state, the number of possible actions are $MN$. 

The agent sequentially modifies the blocks in the image for the fixed number of steps, $T_{max}$. If the agent succeeds in changing the original label of the image within $T_{max}$ modifications, we provide a positive reward of 10. If it fails to modify the label, it receives a negative reward of -1. For each modification, we penalize it with -0.1 reward. Our reward strategy is as follows. 
\[
    r_{t} = 
\begin{cases}
    
    10,& \text{if} \hspace{0.1cm} \arg \max p_{t+1} \neq \arg \max p_t \hspace{0.1cm} \text{and} \hspace{0.1cm} t \leq T_{max}\\
    -1,& \text{if} \hspace{0.1cm} \arg \max p_{t+1} = \arg \max p_t \hspace{0.1cm} \text{and} \hspace{0.1cm} t = T_{max}  \\
    -0.1 & \text{if} \hspace{0.1cm} t \leq T_{max}  \\

\end{cases}
\]
We also experimented with the clipped rewards, $+1$ for success and $-1$ for failure. However, we obtained better results with reward strategy mentioned above. In experimental results, we demonstrate the success rate comparison when the maximum reward ($R_{max}$) is set to 10 and 1.

The goal of the agent is to find the attack policy $\pi$ 
that maximizes the Q value function i.e. expected sum of discounted rewards starting is state $s$, taking an action $a$ and following the policy $\pi$, thereafter. The optimal Q-function follows the Bellman equation
\begin{eqnarray}\label{eq:q}
Q^*(s_t,a_t) = r_t + \gamma \max_{a_{ij}} Q^*(s_{t+1},a_{ij}; \theta)
\end{eqnarray}
where $\gamma \in [0, 1]$ is a discount factor and $\theta$ indicates the parameters of Q function.
We use DQN to learn the optimal Q value function \cite{mnih2013playing} where the Q-values for state, action pairs are estimated by minimizing the Mean Square Error (MSE) loss between current Q estimate and target Q values obtained from Eq. \ref{eq:q}.
We use experience replay memory to store the history of state transitions and rewards. We sample mini-batches from the experience replay, calculate the target Q values and update the DQN. Fig. \ref{fig:inpt117}(a) shows the structure of our DQN.
Input image ($x_t$) is represented by a CNN while its class probability vector ($p_t$) is represented by a DNN. The CNN and DNN representations are concatenated and projected onto the final layer of the size equal to the number of actions i.e. $MN$.

\section{Experimental results}
We experimented with target models trained on MNIST and CIFAR-10 datasets. Early results on Imagenet model are included in the appendix.

For MNIST and CIFAR-10, the CNN model in DQN consists of 2 convolution layers (each with 64 filters of size $5 \times 5$ and stride 1), followed by a dense layer of dimension 128. All the layer have ReLu activation and we also use Dropout (0.5) after each layer. 
The DNN model consists of 1 layer of dimension 128 with ReLu activation. The final layer has a linear activation. The DQN for MNIST and CIFAR-10 has approx. 6M and 8M parameters, respectively. We use the Adam optimizer with learning rate $10^{-3}$. For all our experiments, the discount factor $\gamma$ is set to 0.9, the experience replay memory of size is set to 1000 and the maximum number of blocks which the agent can modify ($T_{max}$) is set to 15. We use a $\epsilon$ - greedy method to control the exploration and exploitation. The code is available on github:   \url{https://github.com/mandareln/deep-q-learning-adversarial}.

\subsection{MNIST dataset}
The MNIST dataset consists of handwritten digits of size $28 \times 28$ . The MNIST target model has approximately 184k parameters and test accuracy of 99.58\%. 
We use blocks of size $2 \times 2$, hence the number possible actions are 196. 
The MNIST test set consist of 10k images. The DQN is trained on randomly chosen 5k images from the test set. Initially, $\epsilon$ is set to 0.9 and it is reduced by 0.1 after every 300 images. Hence, at start, the agent explores the environment by randomly modifying blocks of the image and later exploits the learning to act optimally. 

\begin{figure} [!t]
\centering
\begin{tabular}{c c}

\includegraphics[width=250pt, height = 120pt]{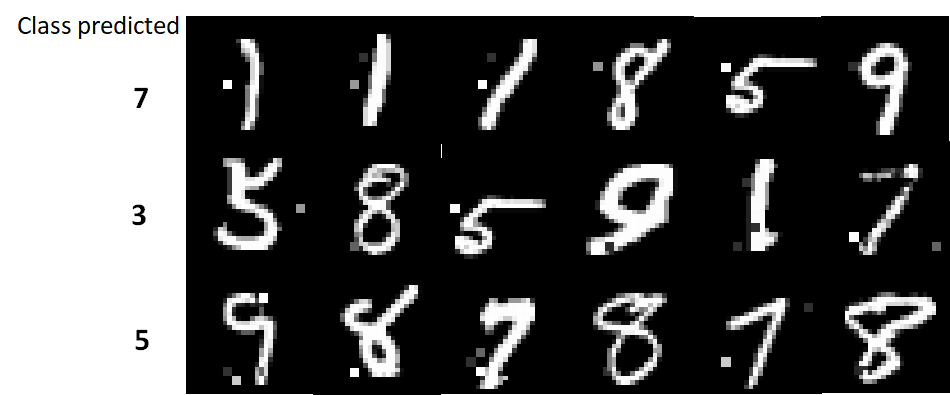}&
\includegraphics[width=170pt, height = 120pt]{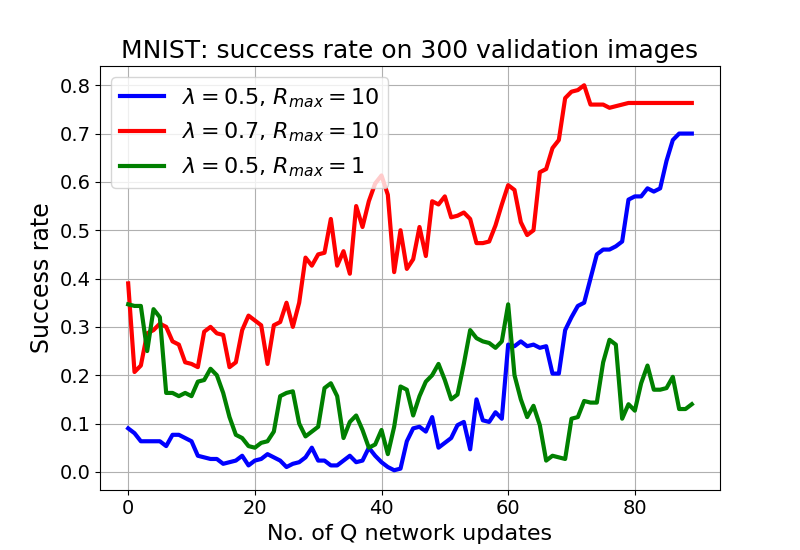}\\(a)&(b)\\
\includegraphics[width=250pt, height = 120pt]{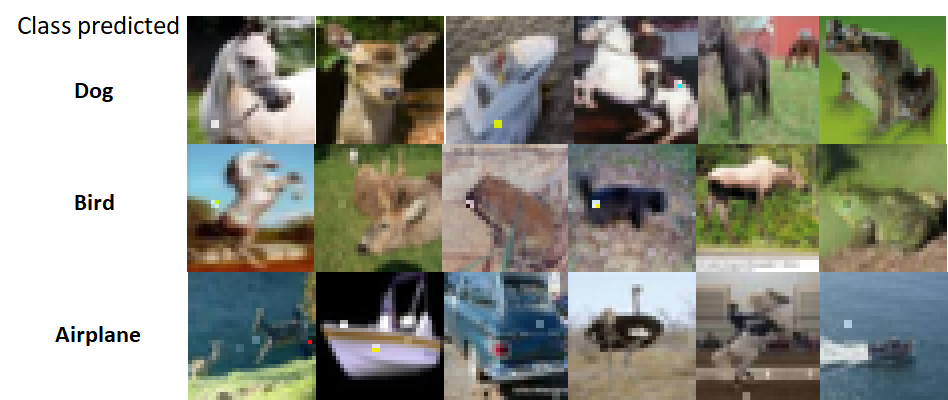}&
\includegraphics[width=170pt, height = 120pt]{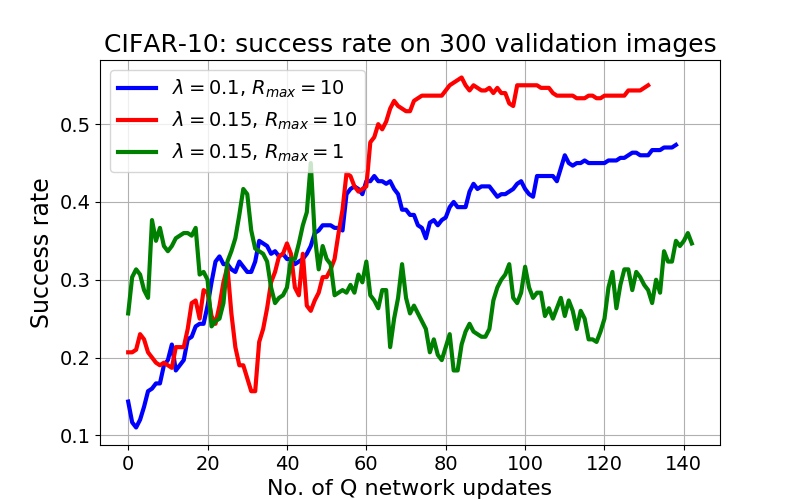}\\(c)&(d)\\

\end{tabular}
\caption{\label{fig:inpt118} Results of RL adversarial attacks (a) images mis-classified by MNIST target where $\lambda = 0.4$  (b) success rate on MNIST validation set over Q network updates (c) images  mis-classified by CIFAR-10 target where $\lambda = 0.05$ (d) success rate on CIFAR validation set over Q network updates.}
\end{figure}

For each image, we store the state transitions and rewards in experience replay, sample a minibatch (of size 32) from it and update the Q network.
To verify the improvement in the attack policy, we use 300 images ( not used in training) as the validation set. After each Q network update, we measure the success rate of the attack on the validation set. The success rate is defined as the percentage of image labels changed when the learned policy is followed. Fig. \ref{fig:inpt118}(b) shows the success rate on validation set for different $\lambda$ . Note that, the agent is able to achieve better success rates over time. We also compared the success rates for $R_{max} = 10$ and $R_{max} = 1$. We achieve better success rate with the higher positive reward.

After training the agent, we execute the learned policy on the remaining 5k samples from the test set and measured the success rate. Table \ref{table:cnn1} shows the success rate comparison for different $\lambda$. With larger value of $\lambda$, a higher success rate is achieved at the cost of making relatively large intensity changes to the image.
As a sanity check, we also measured the success rate of the random attack policy. Results clearly indicates the effectiveness of the learned policy in generating adversarial examples.

Fig. \ref{fig:inpt118}(a) shows the sample images mis-classified by the MNIST target. For most of the images, the agent modifies only couple of blocks which eventually changes the label of the image.
Interestingly, the agent has learned that, to change the label of the digit 1 with minimal modifications, it only has to modify blocks to the left of it. Target model then mis-classifies it as digit 7. Intuitively, after convolutions and pooling layers, modified image of digit 1 would appear as the digit 7 at the classification stage of the target model.

\begin{table}[h]
  \small
  \begin{tabular}{lllllllssslsss}
    \toprule
    \multirow{2}{*}{Policy} & 
      %\vspace{0.5cm}        
      
      \multicolumn{3}{c}{MNIST success rate} & \multicolumn{3}{c}{CIFAR-10 success rate} \\
                 
       & $\lambda = 0.5$ & $\lambda = 0.7$ & $\lambda = 0.8$ & $\lambda = 0.05$ & $\lambda = 0.1 $ & $\lambda = 0.15$ \\
      
      \midrule 
    Learned  & \textbf{0.529} & \textbf{0.726} & \textbf{0.913} & \textbf{0.21} & \textbf{0.439}  & \textbf{0.532} \\
    Random  & 0.006 & 0.021 & 0.044 & 0.052 & 0.109 & 0.166 \\
    \bottomrule
  \end{tabular}
  \caption{\label{table:cnn1} Success rate with learned and random policy for MNIST and CIFAR-10 datasets. }
\end{table} 

\vspace{-0.2cm}

\subsection{CIFAR-10 dataset}
The CIFAR-10 dataset consists of color images of size $32 \times 32$. We use a pre-trained target network which has approximately 776k parameters and test accuracy of 82.16\%. For CIFAR-10 as well, we use blocks of size  $2 \times 2$. Hence, the possible number of actions are 256. 
We performed experiments similar to MNIST, where we use randomly chosen 5k test images to train the DQN and observe the success rate on 300 validation images. Fig. \ref{fig:inpt118}(d) shows the validation performance for $\lambda = 0.1$ and $\lambda = 0.15$ with $R_{max}= 10$. The attack policy improves with more updates. The success rate plot of $R_{max} = 1$ re-iterate the utility of larger positive rewards. Fig. \ref{fig:inpt118}(c) shows the adversarial images generated for $\lambda = 0.05$.

\section{Conclusion}
In this paper, we proposed the deep Q learning based approach which was able to generate adversarial examples with minimal changes to input images. We assumed a semi black-box setting where the RL agent use input image and its class probabilities as state variables. The optimal Q value function is learned by updating the DQN. We observed that the agent eventually learned the attack policy where it was able to locate adversarial sensitive image regions. Experimental results on MNIST, CIFAR-10 and Imagenet datasets demonstrated the effectiveness of our approach.

\bibliographystyle{IEEEtran}
\bibliography{adversial}

\section{Appendix}

\subsection{Results on Imagenet dataset}
\begin{figure} [!h]
\centering
\begin{tabular}{c c}

\includegraphics[width=200pt, height = 130pt]{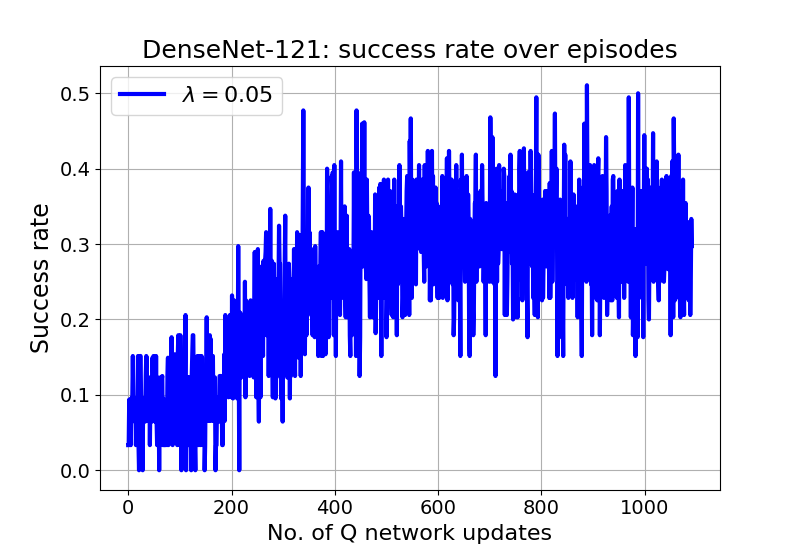}&
\includegraphics[width=200pt, height = 130pt]{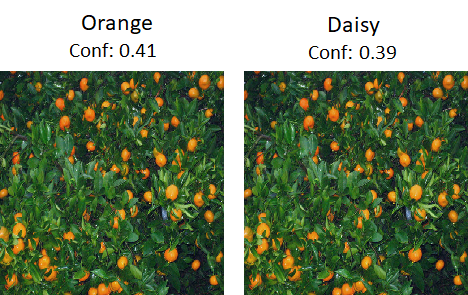}\\(a)&(b)\\
\includegraphics[width=200pt, height = 130pt]{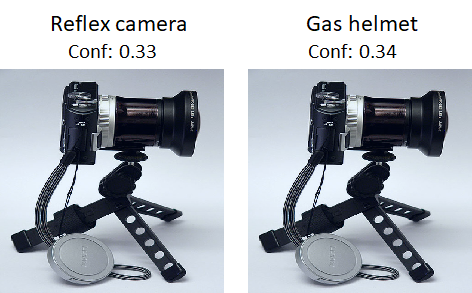}&
\includegraphics[width=200pt, height = 130pt]{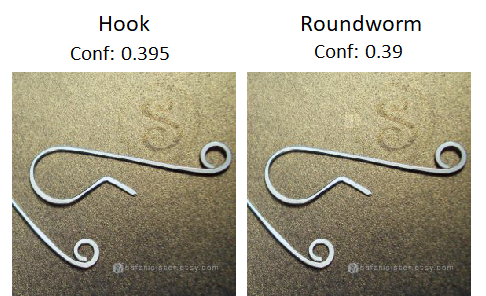}\\(c)&(d)\\

\end{tabular}
\caption{\label{fig:inpt120} Results for DenseNet-121 model (a) success rate over training episodes, (b-d) adversarial images generated by our approach.}
\end{figure}

For the Imagenet dataset, we use DenseNet-121 \cite{huang2017densely} as the target model. The input image size is $224 \times 224$ and we use a blocks of size $14 \times 14$. Hence, the possible number of actions are 256. The DQN model architectures is similar to MNIST/CIFAR-10 except, for the 2 convolution layers in CNN, we use the stride of 2 (instead of 1) in order to reduce the number trainable parameters. The number of parameters of our DQN are approx. 26M.
We trained the DQN model on 70k images from the Imagenet training set. The $\lambda$ value is set to 0.05.

Fig. \ref{fig:inpt120}(a) shows the success rate of the RL agent over the episodes. Fig. \ref{fig:inpt120}(b-d) shows the adversarial images generated by our approach. 
In few of the images, we observed an interesting trend. Whenever possible, the agent attacks a portion of the image such that the modified class is an object/pattern present in the same image e.g. Fig. \ref{fig:inpt120}(d), the original image was classified as Hook and after the attack, the image was classified as worm. We can notice a worm like pattern in the top right potion of the image, where the RL agent has made a slight modification.
Probably, in such cases, the attack executed by the RL agent is altering the attribution of pixels for the classification. Sundararajan et al. \cite{sundararajan2017axiomatic} proposed the use of input gradients for estimating the pixel attribution. 
Though, we do not assume the availability of the model for learning the attack policy, we further plan to investigate this effect using input gradients.

\end{document}